\documentclass[lettersize,journal]{IEEEtran}
\usepackage{amsmath,amsfonts}
\usepackage{algorithmic}
\usepackage{algorithm}
\usepackage{array}
\usepackage[caption=false,font=normalsize,labelfont=sf,textfont=sf]{subfig}
\usepackage{textcomp}
\usepackage{stfloats}
\usepackage{url}
\usepackage{verbatim}
\usepackage{booktabs}
\usepackage{multirow}%
\usepackage{graphicx}
\usepackage{cite}
\usepackage{pgfplots}
\pgfplotsset{compat=1.17}

\begin{document}

\title{Dual Causal Inference: Integrating Backdoor Adjustment and Instrumental Variable Learning for Medical VQA}

\author{Zibo Xu, Qiang Li, Ke Lu, Jin Wang, Weizhi Nie and Yuting Su
\thanks{This work was supported in part by the National Natural Science Foundation of China under Grants U21B2024 and 62272337. \textit{(Corresponding author: Weizhi Nie)}}
\thanks{Zibo Xu and Qiang Li are with the School of Microelectronics, Tianjin University, Tianjin 300072, China (e-mail: xzb6666@tju.edu.cn, liqiang@tju.edu.cn).}
\thanks{Ke Lu is with the Department of Orthopedics, Affiliated Kunshan Hospital of Jiangsu University, Jiangsu 215300, China (e-mail: sgu8434@sina.com).}
\thanks{Jin Wang is with the Department of Clinical Laboratory, The Third Central Hospital of Tianjin, Tianjin 300170, China (e-mail: wangjane@126.com).}
\thanks{Weizhi Nie and Yuting Su are with the School of Electrical and Information Engineering, Tianjin University, Tianjin 300072, China (e-mail: weizhinie@tju.edu.cn, ytsu@tju.edu.cn). }}



\maketitle

\begin{abstract}

Medical Visual Question Answering (MedVQA) aims to generate clinically reliable answers conditioned on complex medical images and questions. However, existing methods often overfit to superficial cross-modal correlations, neglecting the intrinsic biases embedded in multimodal medical data. Consequently, models become vulnerable to cross-modal confounding effects, severely hindering their ability to provide trustworthy diagnostic reasoning. To address this limitation, we propose a novel \textit{Dual Causal Inference} (DCI) framework for MedVQA. To the best of our knowledge, DCI is the first unified architecture that integrates Backdoor Adjustment (BDA) and Instrumental Variable (IV) learning to jointly tackle both observable and unobserved confounders. Specifically, we formulate a Structural Causal Model (SCM) where observable cross-modal biases (e.g., frequent visual and textual co-occurrences) are mitigated via BDA, while unobserved confounders are compensated using an IV learned from a shared latent space. To guarantee the validity of the IV, we design mutual information constraints that maximize its dependence on the fused multimodal representations while minimizing its associations with the unobserved confounders and target answers. Through this dual mechanism, DCI extracts deconfounded representations that capture genuine causal relationships. Extensive experiments on four benchmark datasets—SLAKE, SLAKE-CP, VQA-RAD, and PathVQA—demonstrate that our method consistently outperforms existing approaches, particularly in out-of-distribution (OOD) generalization. Furthermore, qualitative analyses confirm that DCI significantly enhances the interpretability and robustness of cross-modal reasoning by explicitly disentangling true causal effects from spurious cross-modal shortcuts.

\end{abstract}


\section{Introduction}

\label{sec:introduction}

In recent years, with the rapid development of visual question answering (VQA)~\cite{Lu2023MultiscaleFE,guo2023from, vosoughi2024cross}, Medical Visual Question Answering (MedVQA)~\cite{wu2023pmcllama,zhang2023pmcvqa,eslami2023pubmedclip,li2023masked,chen2022m3ae} has emerged as an important research direction connecting computer vision, natural language processing, and medical image analysis. Different from general-domain VQA, MedVQA tasks are highly specialized. They require models to accurately comprehend both medical images and domain-specific clinical questions to provide reliable answers for clinical decision-making. With the increasing complexity of modern healthcare needs, developing accurate and robust MedVQA systems has become indispensable~\cite{li2023llavamed}.

Despite these advances, existing MedVQA methods still face significant challenges, primarily due to widespread data biases in medical datasets. These biases stem from patient differences, imbalanced disease distributions, varying imaging devices, and the diverse ways doctors write clinical reports, thereby injecting non-causal statistical priors into the training distributions.~\cite{Nie2023ChestXI}. Consequently, models often exploit spurious correlations between images and questions rather than learning genuine causal relationships~\cite{Liu2022CrossModalCR,LuKnowledge, weninference}. For example, a model might rely on textual shortcuts (e.g., specific question styles directly leading to certain answers) instead of true medical evidence, severely limiting its reliability in real-world clinical scenarios. Furthermore, many existing causal inference methods treat image and text modalities separately. However, clinical confounders frequently manifest as cross-modal entanglements (e.g., specific visual artifacts strongly co-occurring with certain question templates). This simplification underestimates the complex cross-modal interactions, restricting the model to shallow causal relationships and preventing it from capturing deeper multimodal dependencies.

\begin{figure}[t]
  \centering
   \includegraphics[width=\linewidth]{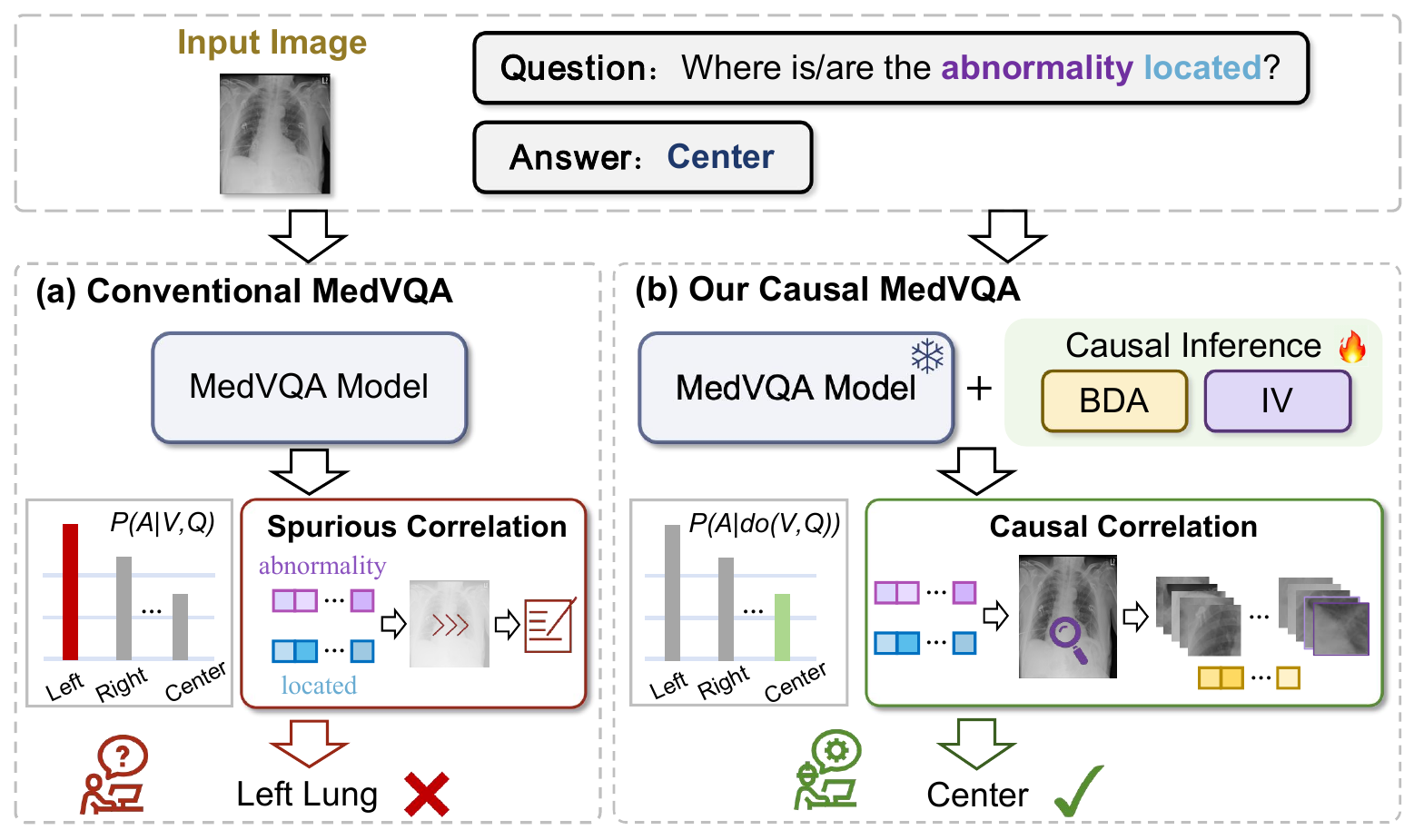}
   \caption{
   \textbf{Motivation for applying causal inference in MedVQA.} (a) Conventional MedVQA models estimate the observational probability $P(A|V,Q)$. They easily fall prey to \textbf{spurious correlations} driven by dataset biases, leading to incorrect predictions. (b) Our proposed framework intervenes via $P(A|do(V,Q))$ utilizing a dual causal approach (BDA and IV). This structurally mitigates confounding effects, enabling the model to capture genuine \textbf{causal correlations} and yield the correct answer.
   }
   \label{fig:fig1}
\end{figure}

To address these limitations, it is necessary to explicitly introduce causal modeling into MedVQA. While causal inference has shown promise in enhancing the interpretability and robustness of models in healthcare~\cite{Pearl2009CausalII,yang2021causal}, current methods mainly rely on heuristic bias mitigation or attention calibration. They often fail to explicitly model the causal dependencies among images, questions, and answers. More importantly, real-world MedVQA is affected by both observable confounders (e.g., explicit phrase frequencies or visual artifacts) and unobserved confounders (e.g., latent patient characteristics or hidden dataset biases). If not properly addressed, these confounders will continuously induce spurious cross-modal associations, misleading the reasoning process.

To provide an intuitive illustration of our motivation, Fig.~\ref{fig:fig1} contrasts a conventional MedVQA model with our proposed causal inference framework. As shown in Fig.~\ref{fig:fig1}(a), existing models often overfit to superficial statistical correlations instead of performing genuine multimodal reasoning. For instance, the model may form a spurious language-only shortcut, relying solely on specific textual tokens (e.g., ``abnormality'' and ``located'') to guess the answer, completely bypassing the visual evidence. This shortcut reflects how confounders distort the learned representations, leading to diagnostic hallucinations. In contrast, Fig.~\ref{fig:fig1}(b) illustrates our approach, where textual semantics and visual features are jointly aligned and purified. By explicitly blocking the confounding pathways, the model moves beyond surface correlations to establish faithful cross-modal groundings, successfully identifying the correct visual region and yielding the accurate answer.

Specifically, we propose a novel and unified causal learning framework termed \textbf{\textit{Dual Causal Inference (DCI)}} for MedVQA. To the best of our knowledge, DCI is the \textbf{first} work to jointly introduce \textbf{\textit{Backdoor Adjustment (BDA)}} and \textbf{\textit{Instrumental Variable (IV)}} estimation into a single, end-to-end trainable architecture for multimodal reasoning. The core idea of DCI is to systematically neutralize the dual threats of observable and unobserved confounders through these two complementary causal mechanisms. The BDA module targets observable confounders, such as the frequent co-occurrences of certain pathologies, anatomical regions, or biased question templates in the training data. BDA constructs modality-specific confounder dictionaries and performs causal intervention by stratifying these observable biases. This operation explicitly breaks the spurious associations between visual and linguistic inputs, ensuring that the fused multimodal representations capture true, unbiased causal evidence.
Simultaneously, the IV module focuses on unobserved confounders(e.g., undocumented patient traits or hidden dataset alignment gaps) that cannot be directly measured but still heavily distort causal inference. To neutralize these hidden effects, we introduce a latent instrument I, learned from the same data space as the latent confounder C. By optimizing mutual information objectives, the model maximizes the dependence between I and the multimodal features X, while minimizing the dependence between I and both C and the target answer A. This theoretical constraint ensures that I acts as a valid instrument, influencing A exclusively through X, thereby achieving unbiased causal estimation even in the presence of unobserved confounders.

By integrating BDA and IV into a cohesive framework, DCI produces deconfounded multimodal representations that faithfully reflect the genuine causal relationships among medical images, questions, and answers. This dual-level causal defense significantly enhances both the robustness and interpretability of MedVQA, paving the way for more reliable, clinically grounded decision-making.

Our contributions are summarized as follows:
\begin{itemize}
    \item We propose DCI, the first unified causal learning framework for MedVQA that systematically integrates Backdoor Adjustment and Instrumental Variable Learning. This dual-defense strategy effectively neutralizes the risks of both observable and unobserved confounders, preventing the model from learning superficial cross-modal shortcuts.

    \item We design a novel mutual information-driven mechanism for learning latent instrumental variables. By rigorously maximizing the dependence on fused multimodal representations while minimizing associations with both unobserved confounders and target answers, this mechanism ensures valid causal disentanglement and theoretically grounded unbiased estimation.

    \item We conduct extensive experiments on four benchmark datasets, including SLAKE, SLAKE-CP, VQA-RAD, and PathVQA. Comprehensive quantitative results, ablation studies, and qualitative analyses demonstrate that our DCI framework achieves state-of-the-art performance. Crucially, it significantly enhances out-of-distribution (OOD) generalization and mitigates query-based hallucinations, proving its robustness under complex clinical confounding settings.
\end{itemize}

\section{Related Work}

\subsection{Medical Visual Question Answering}
MedVQA has witnessed remarkable progress, primarily driven by advancements in representation learning and fine-grained cross-modal alignment. A prominent line of research focuses on adapting large language models (LLMs) and foundational vision-language models to the medical domain \cite{wu2023pmcllama, zhang2023pmcvqa, li2023llavamed, Sonsbeek2023OpenEndedMV, eslami2023pubmedclip}. Most notably, Lin \textit{et al.} \cite{lin2023pmc} introduced PMC-CLIP, which aligns biomedical images and text on large-scale datasets, establishing a highly robust backbone for MedVQA tasks. To further overcome annotated data scarcity, various self-supervised and contrastive paradigms have been proposed \cite{Nguyen2019OvercomingDL, Liu2021ContrastivePA, M2I2}. For instance, Li \textit{et al.} \cite{li2023masked} utilized masked modeling combined with cross-modal contrastive objectives, while Chen \textit{et al.} \cite{chen2022m3ae} devised a multimodal masked autoencoder that learns profound domain knowledge by reconstructing missing pixels and tokens. Beyond pre-training, another crucial trajectory centers on enhancing fine-grained cross-modal interaction. AMAM \cite{pan2022amam} and WSDAN \cite{huang2023dual} incorporate dual-attention mechanisms and textual semantic extraction to capture complex intermodal dependencies. Similarly, Tascon-Morales \textit{et al.} \cite{tascon2023localized} introduced a multi-glimpse attention module that integrates global context before attending to specific regions of interest. Building upon these fine-grained alignments, recent efforts like MOTOR \cite{shaaban2025motor} extend MedVQA into a training-free multi-modal retrieval-augmented generation framework to fetch clinically relevant contexts.

\subsection{Causal Inference in Vision-Language Learning}
Causal inference has emerged as a powerful paradigm to address spurious correlations and dataset biases across various vision-language tasks. In the medical domain, causal frameworks have been successfully adopted to improve weakly supervised semantic segmentation \cite{NEURIPS-sseg}. Furthermore, researchers have applied causal interventions to chest X-ray classification by modeling confusing regions as confounders via backdoor adjustment \cite{Nie2023ChestXI} and isolating disease-specific representations \cite{li2025multi}. Chen \textit{et al.} \cite{chen2023cross} introduced a Visual-Linguistic Causal Intervention utilizing front-door modules to disentangle image-text confounders for robust medical report generation. For general VQA, Yang \textit{et al.} \cite{yang2021causal} proposed Causal Attention to explicitly remove elusive confounding effects. Subsequent advanced methods have further fortified causal reasoning by discovering latent cross-modal causal structures \cite{Liu2022CrossModalCR}, integrating multimodal LLMs for collaborative inference \cite{weninference}, and mitigating the inherent bias of answer-heuristic frameworks \cite{LuKnowledge}. Specifically for Medical VQA, overcoming severe modality biases and language priors is critical. DeBCF \cite{zhan2023debiasing} and DeCoCT \cite{decoct} employ counterfactual contrastive training to effectively reduce language priors and question sensitivity. CCIS-MVQA \cite{cai2024counterfactual} extends this by generating counterfactual image samples to construct causal explanations. More recently, CIMB-MVQA \cite{liu2025cimb} performs explicit causal interventions to suppress modality-specific cross-modal biases.

\subsection{Instrumental Variables and Representation Learning}
Instrumental variable methods have recently attracted growing attention in both theoretical and applied machine learning as a means to mitigate unobserved confounding and improve causal representation learning.
Nie et al.~\cite{Nie2023InstrumentalVL} proposed an Instrumental Variable Learning framework for chest X-ray classification, which combines IV estimation with medical semantic fusion to eliminate confounding effects, thus improving the interpretability and reliability of diagnostic predictions.
From a theoretical perspective, Yuan et al.~\cite{yuan2023instrumental} introduced IV-DG, which is a domain generalization method. It uses causal graphs to distinguish between domain-invariant and domain-specific components, and uses the features of one domain as an instrument for another to remove biases from unobserved confounders and improve model generalization.
Wu et al.~\cite{wu2025instrumental} provided a comprehensive survey of instrumental variables in causal inference and machine learning, reviewed identification conditions, estimation algorithms and IV synthesis methods.
Pan et al.~\cite{pan2024unbiased} proposed leveraging instrumental variables to enhance the causal effect between visual features and answers, thereby mitigating spurious correlations.

\section{METHODOLOGY}

\subsection{Causal Graph of MedVQA}
\begin{figure}
  \centering
   \includegraphics[width=\linewidth]{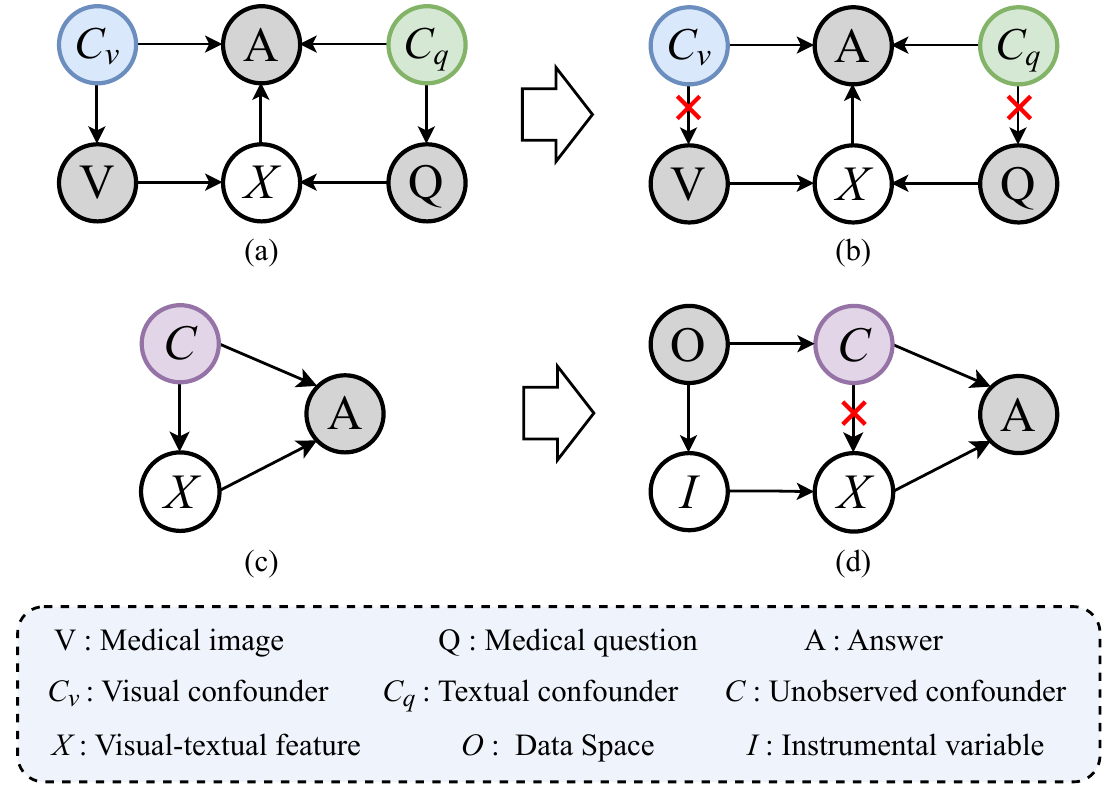}
   \caption{
   Structural Causal Model (SCM) for multi-modal MedVQA. Gray nodes denote observed variables. \textbf{(a)} Observable confounders ($C_v, C_q$) introduce spurious cross-modal correlations. \textbf{(b)} BDA severs these visible backdoor paths (red crosses). \textbf{(c)} An unobserved confounder ($C$) creates a latent spurious path ($X \leftarrow C \rightarrow A$) between multi-modal features and answers. \textbf{(d)} IV learning synthesizes an instrument $I$ to cut the latent confounding link, isolating the genuine causal effect.
   }
   \label{fig:causal}
\end{figure}

We formulate the multi-modal MedVQA process using a Structural Causal Model (SCM), as shown in Fig.~\ref{fig:causal}. The graph explicitly encodes how observable and unobserved confounders affect the relationship between medical images, questions, and answers~\cite{NiPS-longtail,NEURIPS-sseg,Pearl2009CausalII}.

\noindent \textbf{Observable Confounders \& Backdoor Paths.}
Ideally, the reasoning follows the pure causal path $\{V, Q\} \to X \to A$, where the multi-modal fused feature $X$ dictates the answer. However, as shown in Fig.~\ref{fig:causal}(a), observable confounders $C_v$ (e.g., specific imaging protocols) and $C_q$ (e.g., linguistic priors) introduce spurious cross-modal correlations. They establish backdoor paths $V \leftarrow C_v \to A$ and $Q \leftarrow C_q \to A$, misleading the model to rely on superficial co-occurrences. To achieve the true causal intervention $P(A|do(V,Q))$ as depicted in Fig.~\ref{fig:causal}(b), we must cut these paths using BDA.

\noindent \textbf{Latent Confounders \& Instrumental Variables.}
Beyond observable biases, MedVQA is further plagued by unobserved confounders $C$, which create a latent backdoor path $X \leftarrow C \to A$ (Fig.~\ref{fig:causal}(c)). Since $C$ cannot be directly marginalized, we introduce an instrumental variable $I$ synthesized from the observable data space $O$ (Fig.~\ref{fig:causal}(d)). A valid instrument $I$ strictly drives $X$ ($I \to X$) but remains independent of $C$, influencing $A$ only through its effect on $X$. By optimizing these structural conditions via mutual information constraints, $I$ enables the causal disentanglement of multi-modal features, effectively neutralizing the hidden effects of $C$.

\begin{figure*}[!ht]
 \centering
	\includegraphics[width=0.9\textwidth]{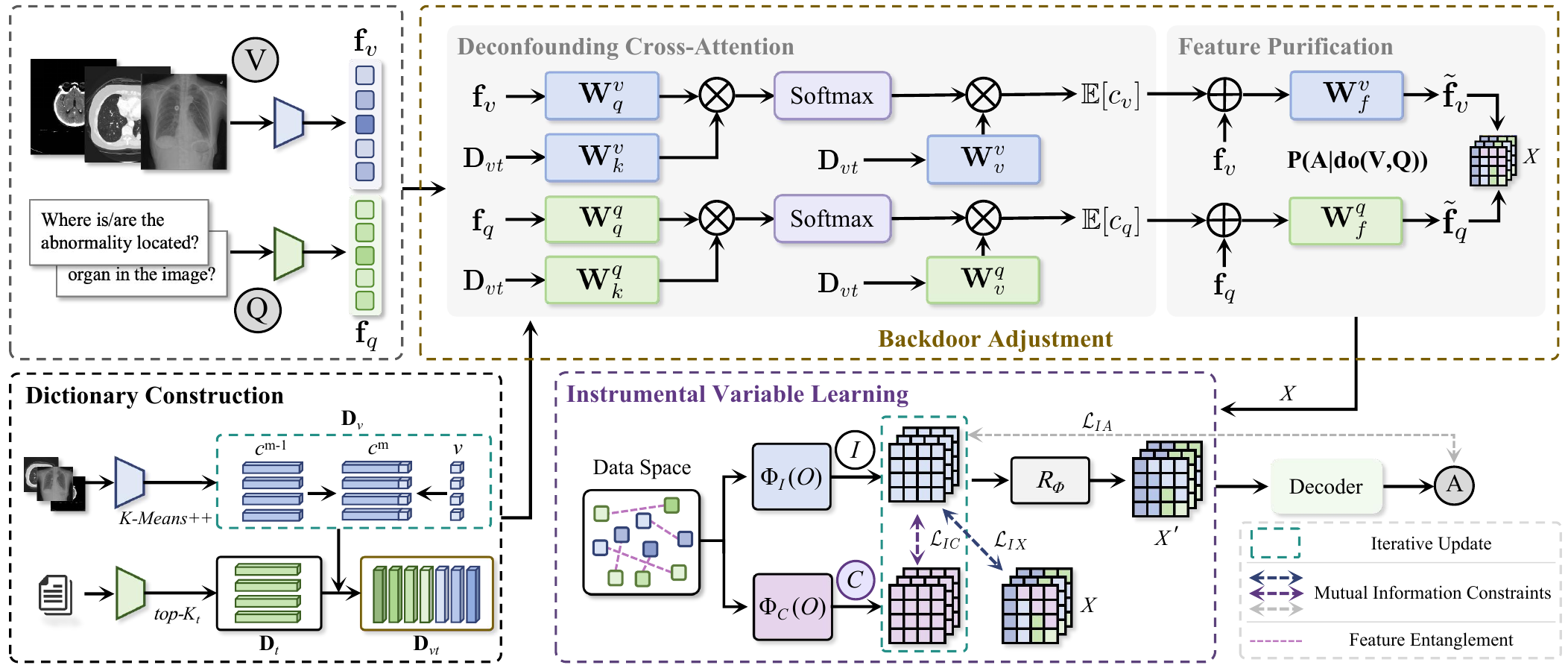}
	\caption{\textbf{Overview of the proposed Dual Causal Inference (DCI) framework.} It employs a sequential deconfounding paradigm to mitigate both observable and unobserved biases. The \textbf{Backdoor Adjustment} module (top) leverages confounder dictionaries to eliminate explicit dataset biases, yielding the observable-deconfounded representation $X$. Subsequently, the \textbf{Instrumental Variable Learning} module (bottom) distills a valid latent instrument $I$ via mutual information constraints to block unmeasured confounders. Finally, a causal regressor $R_{\Phi}$ maps $I$ into the ultimately purified representation $X'$ for the Decoder, ensuring robust medical reasoning..}
    \label{net}
\end{figure*}

\subsection{Backdoor Adjustment for Observable Confounders}
\label{sec:bda}
The primary goal of causal intervention in MedVQA is to eliminate the spurious dependencies introduced by observable confounders, thereby enabling the model to focus on genuine causal relationships between the multimodal inputs and the medical answers.
As illustrated in our SCM(Fig.~\ref{fig:causal}(b)), there exist two primary backdoor paths: $V \leftarrow C_v \rightarrow A$ and $Q \leftarrow C_q \rightarrow A$.
These paths introduce statistical correlations that do not reflect the true causal influence, but rather the observational biases inherent in the dataset.

From a probabilistic perspective, the observed prediction process in conventional MedVQA models is governed by Bayes' theorem, which can be represented as:
\begin{equation}
\label{eq:obs_prob}
P(A \mid V, Q) = \sum_{c_v \in \mathcal{C}_v} \sum_{c_q \in \mathcal{C}_q} P(A \mid V, Q, c_v, c_q)P(c_v, c_q \mid V, Q),
\end{equation}
where $\mathcal{C}_v$ and $\mathcal{C}_q$ denote the spaces of visual and textual confounders, respectively. 
The conditional probability $P(c_v, c_q \mid V, Q)$ quantifies how confounders co-vary with the observed modalities. 
In practice, this term reflects dataset-specific shortcuts. For example, if the dataset contains a disproportionate number of ``Chest X-ray--pneumonia'' pairs, the model maximizes the likelihood by merely associating the frequent visual pattern (e.g., general lung opacities) or question templates with the answer ``pneumonia'', even when such correlation is spurious. 
Hence, $P(c_v, c_q \mid V, Q)$ acts as a confounder prior that distorts multimodal causal inference.

To block these spurious backdoor paths, we apply the \textbf{do-operator}~\cite{Pearl2009AnIT} from causal calculus, which mathematically represents an active intervention by entirely cutting off the incoming edges to the intervened variables $V$ and $Q$.
Instead of relying on passive observations, we explicitly formulate the joint interventional distribution:
\begin{equation}
\begin{aligned}
\label{eq:do_joint}
&P(A \mid \mathrm{do}(V, Q)) =
\\ &\sum_{c_v} \sum_{c_q} P(A \mid \mathrm{do}(V, Q), c_v, c_q) P(c_v, c_q \mid \mathrm{do}(V, Q)).
\end{aligned}
\end{equation}

According to the rules of do-calculus, performing the intervention $\mathrm{do}(V, Q)$ severs the causal links $C_v \rightarrow V$ and $C_q \rightarrow Q$. Consequently, the confounders $C_v$ and $C_q$ are no longer causally influenced by or statistically dependent on the specific instances of $V$ and $Q$. Thus, we have:
\begin{equation}
\label{eq:simplify_do}
P(c_v, c_q \mid \mathrm{do}(V, Q)) = P(c_v, c_q).
\end{equation}
Furthermore, assuming that the modality-specific observable confounders are independently distributed in their marginal spaces prior to multimodal fusion, we can decouple the joint prior as $P(c_v, c_q) \approx P(c_v)P(c_q)$.
Substituting this into Eq.~\eqref{eq:do_joint}, we derive the final tractable form of the BDA:
\begin{equation}
\label{eq:do_final}
P(A \mid \mathrm{do}(V, Q)) \approx \sum_{c_v} \sum_{c_q} P(A \mid V, Q, c_v, c_q) P(c_v) P(c_q).
\end{equation}

This represents the theoretical essence of our BDA module: it replaces the biased, dataset-dependent conditional distribution $P(c_v, c_q\mid V, Q)$ with the universal marginal priors $P(c_v)$ and $P(c_q)$. 
By forcing the model to marginalize over all possible confounders fairly, we conceptually transform the MedVQA reasoning paradigm from correlation-based curve fitting $P(A \mid V, Q)$ to rigorous causal reasoning $P(A \mid \mathrm{do}(V, Q))$. 
This deconfounding process yields a purified multimodal representation, establishing a reliable basis for answer generation driven by true medical semantics rather than statistical coincidences.

\noindent\textbf{Backdoor Adjustment in MedVQA.}
To instantiate the theoretical causal intervention derived above, we seamlessly integrate the BDA mechanism into the MedVQA architecture, as shown in Fig.~\ref{net}. Our objective is to enable the model to infer the answer $A$ from the multimodal inputs $V$ and $Q$, while explicitly neutralizing the spurious influence of the confounders $C_v$ and $C_q$.

According to the joint BDA formula in Eq.~\eqref{eq:do_final}, the interventional probability can be reformulated as the expectation of the network's predictive model over the joint confounder space:
\begin{equation}
P(A \mid \mathrm{do}(V, Q)) = \mathbb{E}_{c_v, c_q}[\mathrm{Softmax}(g(V, Q, c_v, c_q))],
\label{eq:fd2}
\end{equation}
where $g(\cdot)$ represents the causal inference function parameterized by the multimodal network logits, and $\mathbb{E}_{c_v, c_q}$ denotes the expectation with respect to the prior confounder distributions. 
However, computing this exact expectation requires multiple forward passes across all possible confounder configurations, which is computationally prohibitive. To tackle this, we employ the Normalized Weighted Geometric Mean (NWGM)~\cite{pmlr-v37-xuc15}, a rigorous approximation technique widely used in multimodal fusion to move the expectation inside the nonlinear activation functions. By applying NWGM, we approximate Eq.~\eqref{eq:fd2} at the feature level:
\begin{equation}
\begin{aligned}
\mathbb{E}_{c_v, c_q}[\mathrm{Softmax}(g(V,& Q, c_v, c_q))] \approx \\& \mathrm{Softmax}\big(g(V, Q, \mathbb{E}[c_v], \mathbb{E}[c_q])\big).
\label{eq:nwgm_approx}
\end{aligned}
\end{equation}
Eq.~\eqref{eq:nwgm_approx} elegantly implies that the model’s deconfounded inference relies not on repetitive sampling, but on integrating the direct observations ($V, Q$) with the causally marginalized expected confounder representations ($\mathbb{E}[c_v], \mathbb{E}[c_q]$).

To compute these expected confounder features in practice, we leverage the joint multimodal confounder dictionary $\mathbf{D}_{vt} \in \mathbb{R}^{N \times d}$, where $N = K_v + K_t$. We model the prior probability of each confounder as an adaptive attention weight. Specifically, given the extracted visual feature $\mathbf{f}_v \in \mathbb{R}^{d}$ and textual feature $\mathbf{f}_q \in \mathbb{R}^{d}$, the BDA is implemented as a cross-attention mechanism that projects both the instance features and the confounder dictionary into a shared semantic space:
\begin{equation}
\begin{aligned}
\mathbb{E}[c_v] &= \mathrm{Softmax}\left(\frac{(\mathbf{f}_v \mathbf{W}_{q}^v) (\mathbf{D}_{vt} \mathbf{W}_{k}^v)^\top}{\sqrt{d_k}}\right) (\mathbf{D}_{vt} \mathbf{W}_{v}^v), \\
\mathbb{E}[c_q] &= \mathrm{Softmax}\left(\frac{(\mathbf{f}_q \mathbf{W}_{q}^q) (\mathbf{D}_{vt} \mathbf{W}_{k}^q)^\top}{\sqrt{d_k}}\right) (\mathbf{D}_{vt} \mathbf{W}_{v}^q),
\end{aligned}
\label{eq:attention_expectation}
\end{equation}
where $\mathbf{W}_{q}^{(\cdot)}$, $\mathbf{W}_{k}^{(\cdot)}$, and $\mathbf{W}_{v}^{(\cdot)}$ are learnable linear projection matrices, and $d_k$ is the scaling factor. 
Through this mechanism, the dictionary $\mathbf{D}_{vt}$ provides discrete support for the possible values of the unobserved confounders, while the attention scores act as a learnable estimate of their marginal prior probabilities $P(c_v)$ and $P(c_q)$.

Finally, to inject these estimated confounder expectations back into the mainstream representation, the original features are concatenated with their corresponding expected confounders and projected via a linear transformation:
\begin{equation}
\begin{aligned}
\tilde{\mathbf{f}}_v = \mathbf{W}_{f}^v [ \mathbf{f}_v \parallel \mathbb{E}[c_v] ], 
\tilde{\mathbf{f}}_q = \mathbf{W}_{f}^q [ \mathbf{f}_q \parallel \mathbb{E}[c_q] ],
\end{aligned}
\label{eq:f_fuse}
\end{equation}
where $[\cdot \parallel \cdot]$ denotes concatenation, and $\mathbf{W}_{f}^{(\cdot)}$ are the fusion weight matrices. Functionally, the adjusted features $\tilde{\mathbf{f}}_v$ and $\tilde{\mathbf{f}}_q$ are liberated from explicit dataset biases. We concatenate them to form the initial multimodal representation $X$. While $X$ successfully mitigates observable confounding, it remains an endogenous variable inherently susceptible to unmeasured latent clinical confounders $C$, necessitating further deep instrumental purification.

\noindent\textbf{Textual Confounder Dictionary ($D_t$).} To capture the semantic variability of medical questions and model textual confounders (e.g., phrase frequency bias) that frequently entangle with visual priors, we establish a unified cross-modal semantic space. Specifically, we extract the top-$K_t$ most frequent and informative medical concepts (including organs, symptoms, and diagnostic verbs) from the training corpus, excluding meaningless stop words. These concepts are then encoded using the frozen text encoder of PMC-CLIP~\cite{lin2023pmc} to construct the textual confounder dictionary $D_t \in \mathbb{R}^{K_t \times d}$, where $K_t$ is the dictionary size and $d$ is the feature dimension. As PMC-CLIP is pre-trained on large-scale biomedical literature, this initialization ensures rich cross-modal medical semantics.

\noindent\textbf{Dynamic Visual Confounder Dictionary ($D_v$).} 
Unlike explicit textual keywords, visual confounders in medical images (e.g., complex lung opacities or cardiomegaly) are highly variable and often form spurious cross-modal correlations with specific question types. Instead of naively averaging class-level features, we construct the visual dictionary $D_v \in \mathbb{R}^{K_v \times d}$ using representative Region-of-Interest (ROI) features extracted from pathologically meaningful areas. To facilitate scalable confounder refinement, we propose an \textit{incremental dynamic updating} strategy. Initially, we employ the K-Means++ algorithm with PCA on the extracted ROI features across the training set to initialize $K_v$ cluster centers. During training, the confounder dictionary is dynamically updated at the $m$-th iteration. Specifically, for each mini-batch ROI feature $\mathbf{v}$, we update its most similar confounder prototype $\mathbf{c}_k \in D_v$ via a momentum moving average:
\begin{equation}
\mathbf{c}_k^{(m)} = \mu \mathbf{c}_k^{(m-1)} + (1 - \mu) \mathbf{v},
\label{eq:dict_update}
\end{equation}
where $\mu \in [0, 1]$ is a momentum coefficient. This efficiently captures evolving multi-modal interactions without the computational overhead of updating over the entire dataset. Empirically, the visual dictionary size $K_v$ is set to 64.

\noindent\textbf{Joint Multimodal Confounder Space ($D_{vt}$).} 
Given the severe cross-modal entanglement of visual pathologies and textual semantics in clinical scenarios, computing interventions within a single modality fails to capture complex multi-modal co-occurrence biases. Therefore, we concatenate the dynamically updated $D_v$ and the static $D_t$ to form a joint visual-textual confounder dictionary $D_{vt} = [D_v \parallel D_t] \in \mathbb{R}^{(K_v + K_t) \times d}$. 
It is worth noting that we adopt an asymmetric updating strategy. Medical textual concepts possess inherently stable semantics; thus, freezing $D_t$ prevents semantic drift and serves as a robust anchor. Conversely, the dynamic $D_v$ adapts to the highly variable visual artifacts. 
By treating $D_{vt}$ as the explicit multi-modal confounder space, the subsequent BDA module can adaptively reweight these cross-modal confounding factors via an attention mechanism, thereby enabling a theoretically grounded multi-modal approximation of the \textit{do}-operator.

\subsection{Instrumental Variable for Unobserved Confounders} 

While BDA mitigates observable biases, unobserved confounders $C$ (e.g., implicit alignment noise) introduce a spurious path $X \leftarrow C \rightarrow A$ (Fig.~\ref{fig:causal}c). Since $C$ is unmeasured, the \textit{do}-calculus in Section~\ref{sec:bda} is inapplicable. To sever this path, we introduce a latent IV, $I$ (Fig.~\ref{fig:causal}d).

A valid instrument $I$ requires~\cite{Pearl2009AnIT,Pearl2009CausalII}: \textbf{(i) Relevance} ($I \not\!\perp\!\!\!\perp X$): inducing exogenous variation in the fused feature $X$; \textbf{(ii) Unconfoundedness} ($I \!\perp\!\!\!\perp C$): independence from confounders; and \textbf{(iii) Exclusion Restriction} ($I \!\perp\!\!\!\perp A \mid (X, C)$): influencing $A$ only via $X$.

Lacking physical IVs in observational MedVQA, we synthesize a latent instrument from the multimodal manifold $O$~\cite{Nie2023InstrumentalVL}. However, extracting both $I$ and $C$ from the identical source $O$ risks violating the core unconfoundedness assumption ($I \!\perp\!\!\!\perp C$). To resolve this, we project $O$ into a shared context via a transformer encoder $\mathrm{Enc}_o(\cdot)$, followed by decoupled MLP heads ($\Phi_i, \Phi_c$) to span the initial feature spaces:
\begin{equation} 
\label{eq:oc_extract} 
I = \Phi_{i}(\mathrm{Enc}_o(O)), \quad C = \Phi_{c}(\mathrm{Enc}_o(O)). 
\end{equation}

While the shared backbone extracts fundamental multimodal semantics, the decoupled heads merely initiate functional separation. To prevent $I$ from degenerating into a confounding feature and strictly enforce $I \!\perp\!\!\!\perp C$, these initial embeddings undergo a rigorous orthogonal decoupling via targeted mutual information constraints, detailed next.

\noindent\textbf{Latent IV Synthesis via Mutual Information Bottleneck.}
To satisfy the three IV conditions (Relevance, Unconfoundedness, and Exclusion), we introduce a Mutual Information (MI) bottleneck. Since exact MI computation in continuous spaces is intractable, we optimize its variational bounds.

\textbf{1) Enforcing Unconfoundedness ($I \perp\!\!\!\perp C$):} To ensure the instrument $I$ is strictly independent of the unobserved confounder $C$, we minimize their MI using the Contrastive Log-ratio Upper Bound (CLUB)~\cite{cheng2020club}. For a mini-batch of size $N$, the unconfoundedness loss $\mathcal{L}_{IC}$ is defined as:
\begin{equation}
\label{mut-IC}
    \mathcal{L}_{IC} = \frac{1}{N^2}\sum_{i=1}^N\sum_{j=1}^N \Big[ \log q_{\theta_{CI}}(C_i \mid I_i) - \log q_{\theta_{CI}}(C_j \mid I_i) \Big],
\end{equation}
where MLP-based $q_{\theta_{CI}}$ approximates $P(C \mid I)$. Indices $i$ and $j$ denote matched positive pairs and unmatched negative samples, respectively. Minimizing this upper bound forces $I$ and $C$ into orthogonal spaces, rigorously preventing $I$ from degenerating into a confounding feature.

\textbf{2) Enforcing Relevance ($I \not\!\perp\!\!\!\perp X$) and Exclusion ($I \perp\!\!\!\perp A \mid X$):} 
$I$ must strongly influence the fused representation $X$ but have no direct shortcut to the answer $A$. Thus, we jointly \textit{maximize} $\mathcal{I}(I; X)$ (via InfoNCE lower bound) and \textit{minimize} $\mathcal{I}(I; A)$ (via CLUB upper bound):
\begin{equation}
\label{mut-IX-IA}
    \begin{aligned}
        \mathcal{L}_{IX} &= - \frac{1}{N^2}\sum_{i=1}^N\sum_{j=1}^N \Big[ \log q_{\theta_{IX}}(X_i \mid I_i) - \log q_{\theta_{IX}}(X_j \mid I_i) \Big],\\
        \mathcal{L}_{IA} &= \frac{1}{N^2}\sum_{i=1}^N\sum_{j=1}^N \Big[ \log q_{\theta_{IA}}(A_i \mid I_i) - \log q_{\theta_{IA}}(A_j \mid I_i) \Big],
    \end{aligned}
\end{equation}
where $q_{\theta_{IX}}$ and $q_{\theta_{IA}}$ are the corresponding estimators, and the negative sign in $\mathcal{L}_{IX}$ enables gradient descent minimization.

Jointly optimizing these bounds actively constructs $I$ into a theoretically valid instrument, rather than relying on natural priors. To derive the ultimate unconfounded representation $X'$, we map the validated instrument $I$ through a causal regressor $R_{\Phi}$. Specifically, $R_{\Phi}$ is instantiated as a three-layer MLP. By generating $X'$ exclusively from the purified instrument $I$ rather than the endogenous $X$, we structurally block any spurious paths stemming from the unobserved confounder $C$. This sequential deconfounding paradigm guarantees robust medical reasoning against unmeasured clinical biases.

\begin{table*}[t]
   \centering
\caption{Comparison of accuracy with SOTA methods on three MedVQA datasets. Questions are categorized into open and closed types. ``Overall'' refers to the accuracy across the entire dataset. The second-best results are underlined.}
\resizebox{0.8\textwidth}{!}
{
\begin{tabular}{lccccccccc}
\toprule
\multicolumn{1}{c}{\multirow{2}{*}{\multirowsetup{\centering}\textbf{Method}}}      & \multicolumn{3}{c}{\textbf{VQA-RAD}} & \multicolumn{3}{c}{\textbf{SLAKE}} & \multicolumn{3}{c}{\textbf{PathVQA}} \\
\cmidrule(lr){2-4} \cmidrule(lr){5-7} \cmidrule(lr){8-10}
            & Open  & Closed & Overall & Open  & Closed & Overall & Open & Closed & Overall \\
\midrule
MEVF-BAN~\cite{Nguyen2019OvercomingDL}    & 49.2  & 77.2  & 66.1 & 77.8  & 79.8  & 78.6 & 8.1 & 81.4 & 44.8 \\
CPRD-BAN~\cite{Liu2021ContrastivePA}    & 52.5  & 77.9  & 67.8 & 79.5  & 83.4  & 81.1 & - & - & - \\
AMAM~\cite{pan2022amam}   & -  & -  & - & 63.8  & 80.3  & 73.3 & 18.2 & 84.4 & 50.4 \\
CLIP-ViT~\cite{Sonsbeek2023OpenEndedMV}   & -  & -  & - & \underline{84.3}  & 82.1  & 83.3 & \underline{40.0} & 87.0 & 63.6 \\
PubMedCLIP~\cite{eslami2023pubmedclip}    & 60.1  & 80.0  & 72.1 & 78.4  & 82.5  & 80.1 & - & - & - \\
M2I2~\cite{M2I2}    & 61.8  & 81.6  & 73.7 & 74.7  & \underline{91.1}  & 81.2 & 36.3 & 88.0 & 62.2 \\
CCIS-MVQA~\cite{cai2024counterfactual}    & 68.8  & 79.2  & 75.1 & 80.1  & 86.7  & 84.1 & - & - & - \\
M3AE~\cite{chen2022m3ae}        & 67.2  & 83.5  & 77.0 & 80.3  & 87.8  & 83.3 & - & - & - \\
PMC-CLIP~\cite{lin2023pmc}    & 67.0  & 84.0  & 77.6 & 81.9  & 88.0  & 84.3 & - & - & - \\
DeCoCT~\cite{decoct}& 67.1 & 85.7 & 78.3 & 82.5 & 87.0  & 84.9 & - & - & - \\
MUMC~\cite{li2023masked}& \underline{71.5} & 84.2 & 79.2 & - & -  & 84.9 & 39.0 & \underline{90.4} & \underline{65.1} \\
CIMB-MVQA~\cite{liu2025cimb}& 69.3 & \underline{86.2} & \underline{79.4} & 82.1 & 89.4  & \underline{85.1} & - & - & - \\
\midrule
\textbf{DCI (Ours)}        & \textbf{74.3}  & \textbf{87.9} & \textbf{82.5} & \textbf{86.8}  & \textbf{91.3} & \textbf{88.6} & \textbf{40.6} & \textbf{92.0} & \textbf{66.4} \\
\bottomrule
\end{tabular}}
\label{table:mainresults}
\end{table*}

\subsection{Training Strategy and Overall Objective}
\label{sec:training}

We optimize our Dual Causal Inference (DCI) framework end-to-end using a joint objective that concurrently addresses task accuracy and causal constraints.

\noindent\textbf{1) Primary VQA Task:} Answer prediction is formulated using standard cross-entropy loss:
\begin{equation}
\label{eq:ce}
\mathcal{L}_{ce} = -\frac{1}{N}\sum_{i=1}^{N} y_i \log \hat{y}_i,
\end{equation}
where $y_i$ and $\hat{y}_i$ are the ground-truth and predicted answer distributions, respectively.

\noindent\textbf{2) Latent IV Constraints:} To guarantee $I$ functions as a valid instrument (satisfying relevance and unconfoundedness), we aggregate the mutual-information bounds:
\begin{equation}
\mathcal{L}_{IV} = \lambda_{1}\mathcal{L}_{IX} + \lambda_{2}\mathcal{L}_{IC} + \lambda_{3}\mathcal{L}_{IA}.
\end{equation}

\noindent\textbf{3) Causal Consistency Constraint:} To suppress observable confounders, we align the biased observational prediction $p_{\theta}(\cdot \mid V, Q)$ (derived from raw features) with the deconfounded causal prediction $p_{\mathrm{do}}(\cdot \mid \mathrm{do}(V, Q))$. The latter is directly computed from the NWGM-adjusted representations ($\textbf{\textit{f}}_{v}^{'}$ and $\textbf{\textit{f}}_{q}^{'}$). We minimize their Kullback-Leibler (KL) divergence to transfer causal invariance:
\begin{equation}
\mathcal{L}_{BD} = \frac{1}{N}\sum_{i=1}^{N} \mathrm{KL}\!\Big( p_{\mathrm{do}}(\cdot \mid \mathrm{do}(V^{(i)},Q^{(i)})) \,\Big\|\, p_{\theta}(\cdot \mid V^{(i)},Q^{(i)}) \Big).
\end{equation}
This regularization compels the network to discard spurious shortcuts in favor of causally sound priors.

\noindent\textbf{Overall Objective:} The final training loss is unified as:
\begin{equation}
\label{eq:total_loss}
\mathcal{L}_{total} = \mathcal{L}_{ce} + \mathcal{L}_{IV} + \lambda_{4}\mathcal{L}_{BD},
\end{equation}
where $\lambda_{1{\sim}4}$ are trade-off hyperparameters. This joint optimization yields representations that are both causally robust and task-effective.

\section{Experiments and Results}
\subsection{Datasets and Implementation Details}
\noindent\textbf{Datasets.}
 VQA-RAD~\cite{Lau2018ADO} is specifically designed for radiology, consisting of 315 images and 3,515 questions with 517 possible answers. Questions are divided into 11 categories: abnormality, attribute, modality, organ system, color, counting, object/condition presence, size, plane, positional reasoning, etc.
SLAKE~\cite{40cb06d16fd1450ea39bfd13d43e9c9f} is a bilingual dataset consisting of 642 images and 14k QA pairs, covering 12 diseases and 39 systemic organs. Diseases mainly include tumors and chest diseases, and the images mainly include the head, neck, chest, and abdomen.
SLAKE-CP~\cite{zhan2023debiasing} is a bias-sensitive variant derived from SLAKE, specifically constructed to evaluate the debiasing capabilities of Med-VQA models. By systematically regrouping and re-splitting the original QA pairs based on question types and answers, it creates disparate prior distributions between the training and testing sets, serving as a rigorous benchmark for out-of-distribution robustness.
PathVQA~\cite{He2020PathVQA3Q} is a pathological image dataset containing 4998 pathological images, and 32,799 QA pairs. Each image is accompanied by multiple questions covering various aspects such as location, shape, color, appearance, etc. We use the official splits across all datasets.

\noindent\textbf{Implementation Details.}
In the training stage, we employ the AdamW optimizer~\cite{adamw} with a batch size of 32, a learning rate of $1\times10^{-4}$, and a weight decay of 0.01. A linear warm-up is applied over the first 5\% of total training steps.The visual and textual encoders are initialized using the pretrained PMC-CLIP~\cite{lin2023pmc} to extract foundational medical features. In Eq.~\eqref{eq:total_loss}, we set $\lambda_1=\lambda_2=0.1$, $\lambda_3=0.05$, and $\lambda_4=0.5$. All models are implemented in PyTorch and trained on two NVIDIA RTX 4090 GPUs.

\subsection{Main Results}

Table~\ref{table:mainresults} summarizes the performance comparison between our method and several state-of-the-art (SOTA) models on three MedVQA datasets: VQA-RAD, SLAKE, and PathVQA. Across all datasets, our method establishes new state-of-the-art results, confirming its strong generalization ability and robust causal reasoning capabilities in multimodal medical tasks.

\noindent\textbf{VQA-RAD:}
On the VQA-RAD dataset, which features complex and descriptive answers, our DCI model achieves an overall accuracy of 82.5\%, outperforming the latest best method (CIMB-MVQA) by 3.1\%. Notably, we observe a significant 2.8\% gain on the challenging open-ended questions compared to MUMC. These results demonstrate that our dual causal adjustment allows the model to bypass superficial dataset biases and focus on genuine multimodal causal links, enabling it to handle long and descriptive reasoning scenarios effectively.

\noindent\textbf{SLAKE:}
The SLAKE dataset is characterized by rich anatomical diversity and concise answers. Here, our method achieves an overall accuracy of 88.6\%, surpassing the strong baseline CIMB-MVQA by 3.5\%. Furthermore, it yields a 2.5\% improvement over CLIP-ViT on open-ended questions. These findings indicate that the DCI framework is highly effective at breaking dataset-specific visual priors. By explicitly modeling and marginalizing the unobserved confounders via latent instruments, the reasoning becomes more precise, even when the provided textual context is limited.

\noindent\textbf{PathVQA:}
PathVQA is the largest and most diverse dataset in our evaluation, focusing on pathology rather than purely macroscopic radiographic images, thus introducing a higher level of visual and semantic variability. Despite this complexity, our approach achieves a 1.3\% overall improvement (66.4\% vs. 65.1\%) and a 1.6\% gain on closed questions over the previous best method, MUMC. This confirms that the proposed feature disentanglement mechanism can seamlessly adapt to complex pathological domains, exhibiting solid robustness across vastly different modalities of medical imaging.

\begin{table*}[t]
\centering
\caption{Ablation study on the core causal components across three MedVQA datasets.}
\label{tab:ablation_main}
\resizebox{0.8\textwidth}{!}{
\begin{tabular}{lccccccccc}
\toprule
\multirow{2}{*}{\textbf{Configuration}} &
\multicolumn{3}{c}{\textbf{VQA-RAD}} &
\multicolumn{3}{c}{\textbf{SLAKE}} &
\multicolumn{3}{c}{\textbf{PathVQA}} \\
\cmidrule(lr){2-4} \cmidrule(lr){5-7} \cmidrule(lr){8-10}
 & Open & Closed & Overall & Open & Closed & Overall & Open & Closed & Overall \\
\midrule
Baseline (no BDA, no IV) & 70.9 & 83.1 & 78.3 & 81.7 & 85.1 & 83.0 & 36.5 & 83.2 & 59.9 \\
w/ BDA only              & 73.7 & 85.7 & 80.9 & 84.5 & 87.7 & 85.8 & 38.6 & 89.1 & 63.9 \\
w/ IV only               & 73.2 & 86.8 & 81.4 & 84.0 & 87.0 & 85.2 & 37.1 & 85.0 & 61.1 \\
\midrule
\textbf{Full Model (DCI)}& \textbf{74.3} & \textbf{87.9} & \textbf{82.5} & \textbf{86.8} & \textbf{91.3} & \textbf{88.6} & \textbf{40.6} & \textbf{92.0} & \textbf{66.4} \\
\bottomrule
\end{tabular}}
\end{table*}

\subsection{Ablation Study}

\begin{table}[t]
\centering
\caption{Detailed ablation on confounder dictionaries and MI loss components.}
\label{tab:ablation_detail}
\resizebox{0.8\columnwidth}{!}{ 
\begin{tabular}{lccc}
\toprule
\textbf{Configuration} & \textbf{VQA-RAD} & \textbf{SLAKE} & \textbf{PathVQA} \\
\midrule
\multicolumn{4}{l}{\textit{(a) Impact of Confounder Dictionaries in BDA}} \\
\midrule
Random $D_v, D_t$ & 76.5 & 83.6 & 59.4 \\
$D_v$ only        & 81.8 & 86.4 & 64.9 \\
$D_t$ only        & 81.4 & 85.5 & 63.6 \\
$D_v + D_t$ (Full)& \textbf{82.5} & \textbf{88.6} & \textbf{66.4} \\
\midrule
\multicolumn{4}{l}{\textit{(b) Impact of MI Loss Components in IV}} \\
\midrule
w/o $\mathcal{L}_{IX}$ & 80.9 & 85.2 & 64.3 \\
w/o $\mathcal{L}_{IC}$ & 82.0 & 86.2 & 64.8 \\
w/o $\mathcal{L}_{IA}$ & 81.8 & 88.1 & 65.1 \\
\bottomrule
\end{tabular}}
\end{table}

\subsubsection{Ablation Study on Core Causal Components}
Table~\ref{tab:ablation_main} details the ablation study evaluating the individual and synergistic contributions of our causal modules. 

\noindent\textbf{Effectiveness of BDA and IV:} 
Integrating the BDA module alone yields substantial overall accuracy improvements over the baseline (+2.6\% on VQA-RAD, +2.8\% on SLAKE, and +4.0\% on PathVQA). This confirms that explicitly marginalizing observable confounders successfully cuts off spurious correlation paths. Similarly, the IV-only variant provides consistent boosts (+3.1\%, +2.2\%, and +1.2\%). By synthesizing a latent instrument and enforcing orthogonal disentanglement, it effectively mitigates unobserved hidden biases (e.g., subjective diagnostic habits) that BDA cannot address.

\noindent\textbf{Synergy of Dual Causal Inference:} 
The full DCI framework (BDA+IV) achieves the highest performance across all metrics. This consistent advantage demonstrates their strong complementarity: BDA handles explicit, observable confounding patterns, while IV neutralizes unmeasured hidden biases. Together, they form a comprehensive dual-defense mechanism that significantly elevates the robustness of multimodal causal reasoning.

\begin{figure}[t]
\centering

\definecolor{mplblue}{RGB}{31, 119, 180}
\definecolor{mplred}{RGB}{214, 39, 40}

\begin{minipage}{0.5\linewidth}
\centering
\begin{tikzpicture}
\begin{axis}[
    width=\linewidth,
    height=5.2cm,
    title={(a) VQA-RAD}, 
    title style={font=\small, yshift=-1mm},
    xlabel={Confounder Dictionary Size $K$},
    ylabel={Overall Accuracy (\%)},
    ymin=74, ymax=84,  
    xmin=0, xmax=9,
    xtick={1, 2, 3.5, 5.5, 8}, 
    xticklabels={16, 32, 64, 128, 256}, 
    grid=major, 
    grid style={solid, gray!40}, 
    legend pos=south east,
    legend style={
        at={(0.5, 0.05)},  
        anchor=south,      
        legend columns=1,  
        legend cell align={left}, 
        draw=gray!50, 
        fill=white, 
        font=\footnotesize,
        rounded corners=1pt
    },
    tick label style={font=\footnotesize},
    label style={font=\footnotesize},
    thick 
]

\addplot[color=mplred, dashed, mark=*, mark size=1.5pt, mark options={solid}] 
    coordinates {(1, 78.7) (2, 80.0) (3.5, 80.9) (5.5, 80.5) (8, 79.8)};
\addlegendentry{w/o IV}

\addplot[color=mplblue, solid, mark=square*, mark size=1.5pt] 
    coordinates {(1, 81.4) (2, 81.8) (3.5, 82.5) (5.5, 82.3) (8, 81.6)};
\addlegendentry{Full DCI}

\end{axis}
\end{tikzpicture}
\end{minipage}%
\hfill
\begin{minipage}{0.5\linewidth}
\centering
\begin{tikzpicture}
\begin{axis}[
    width=\linewidth,
    height=5.2cm,
    title={(b) SLAKE},
    title style={font=\small, yshift=-1mm},
    xlabel={Confounder Dictionary Size $K$},
    ylabel={Overall Accuracy (\%)},
    ymin=80, ymax=90, 
    xmin=0, xmax=9,
    xtick={1, 2, 3.5, 5.5, 8}, 
    xticklabels={16, 32, 64, 128, 256},
    grid=major,
    grid style={solid, gray!40},
    legend pos=south east,
    legend style={
        at={(0.5, 0.05)},  
        anchor=south,  
        legend columns=1, 
        legend cell align={left}, 
        draw=gray!50, 
        fill=white, 
        font=\footnotesize,
        rounded corners=1pt
    },
    tick label style={font=\footnotesize},
    label style={font=\footnotesize},
    thick
]

\addplot[color=mplred, dashed, mark=*, mark size=1.5pt, mark options={solid}] 
    coordinates {(1, 83.7) (2, 85.0) (3.5, 85.8) (5.5, 85.3) (8, 84.9)};
\addlegendentry{w/o IV}

\addplot[color=mplblue, solid, mark=square*, mark size=1.5pt] 
    coordinates {(1, 85.4) (2, 87.2) (3.5, 88.6) (5.5, 88.0) (8, 87.7)};
\addlegendentry{Full DCI}

\end{axis}
\end{tikzpicture}
\end{minipage}

\vspace{-2mm}
\caption{Hyperparameter sensitivity of the dictionary size $K$. 
}
\label{fig:dictionary_size_ablation}
\vspace{-4mm} 
\end{figure}
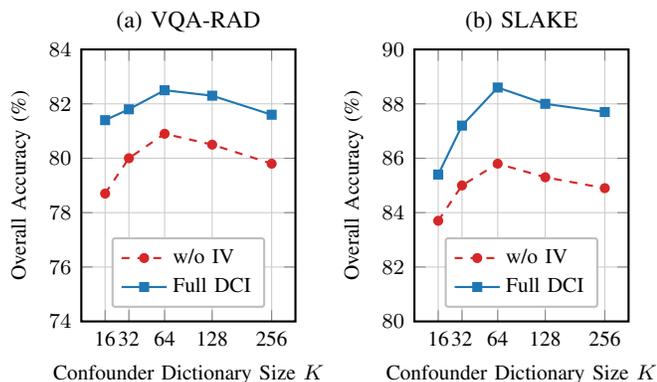

\begin{table}[t]
\centering
\caption{Comparison with SOTA methods on the Out-of-Distribution dataset (SLAKE-CP).}
\label{tab:ood_sota_comparison}
\resizebox{0.8\columnwidth}{!}{ 
\begin{tabular}{lccc}
\toprule
\multirow{2}{*}{\textbf{Methods}} & \multicolumn{3}{c}{\textbf{SLAKE-CP Accuracy (\%)}} \\
\cmidrule(lr){2-4}
 & Open & Closed & Overall \\
\midrule
MEVF+BAN \cite{Nguyen2019OvercomingDL}   & $13.0_{\pm1.4}$ & $29.8_{\pm1.2}$ & $29.1_{\pm1.3}$ \\
PubMedCLIP \cite{eslami2023pubmedclip}    & $13.4_{\pm1.2}$ & $30.5_{\pm1.1}$ & $30.0_{\pm1.2}$ \\
CPRD+BAN \cite{Liu2021ContrastivePA}     & $13.9_{\pm1.3}$ & $31.2_{\pm1.5}$ & $30.4_{\pm1.5}$ \\
DeBCF \cite{zhan2023debiasing}         & $18.6_{\pm1.1}$ & $35.4_{\pm1.0}$ & $34.2_{\pm1.2}$ \\
MISS \cite{chen2024miss}               & $17.3_{\pm1.1}$ & $49.2_{\pm1.0}$ & $33.8_{\pm1.1}$ \\
M2I2 \cite{M2I2}                 & $17.3_{\pm1.1}$ & $51.9_{\pm0.9}$ & $35.2_{\pm1.0}$ \\
M3AE \cite{chen2022m3ae}               & $24.4_{\pm1.0}$ & $65.1_{\pm0.9}$ & $45.4_{\pm1.0}$ \\
DeCoCT \cite{decoct}                   & $27.3_{\pm0.5}$ & $69.6_{\pm0.5}$ & $49.2_{\pm0.6}$ \\
\midrule
\textbf{DCI (Ours)}                    & \textbf{30.2}$_{\pm0.4}$ & \textbf{74.1}$_{\pm0.7}$ & \textbf{52.9}$_{\pm0.6}$ \\
\bottomrule
\end{tabular}
 }
\end{table}

\begin{figure*}[]
 \centering
\includegraphics[width=0.8\linewidth]{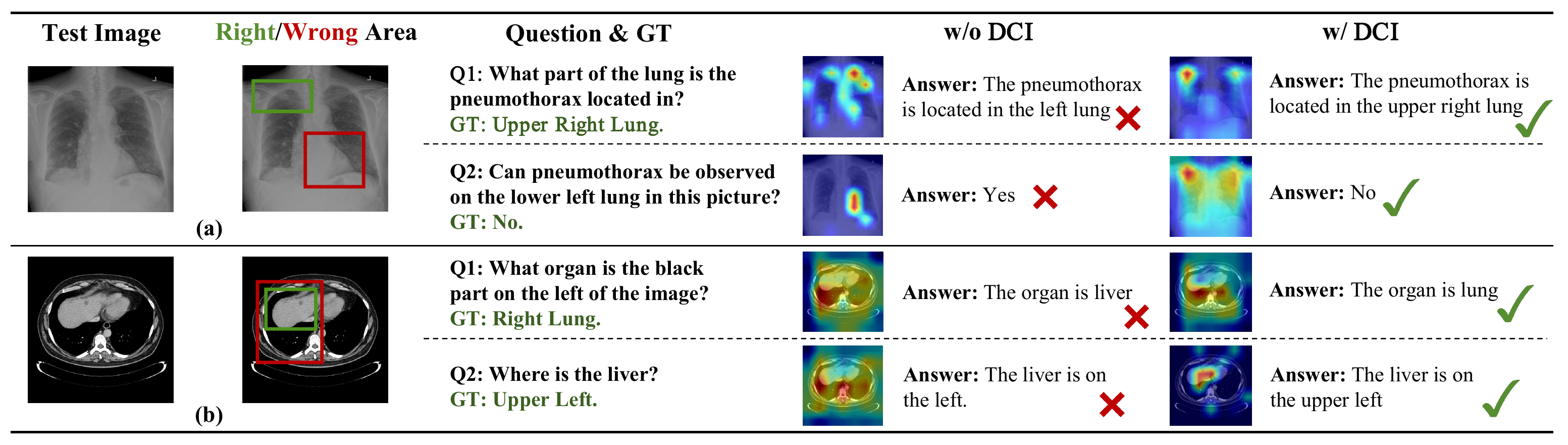}
 \caption{Qualitative results on SLAKE under the challenging ``one-image, multi-question'' scenario. Green bounding boxes denote the true reasons, red bounding boxes denote the misleading areas.
 }
 \label{fig:quaresult}
\end{figure*}

\begin{figure}[]
 \centering
\includegraphics[width=0.8\linewidth]{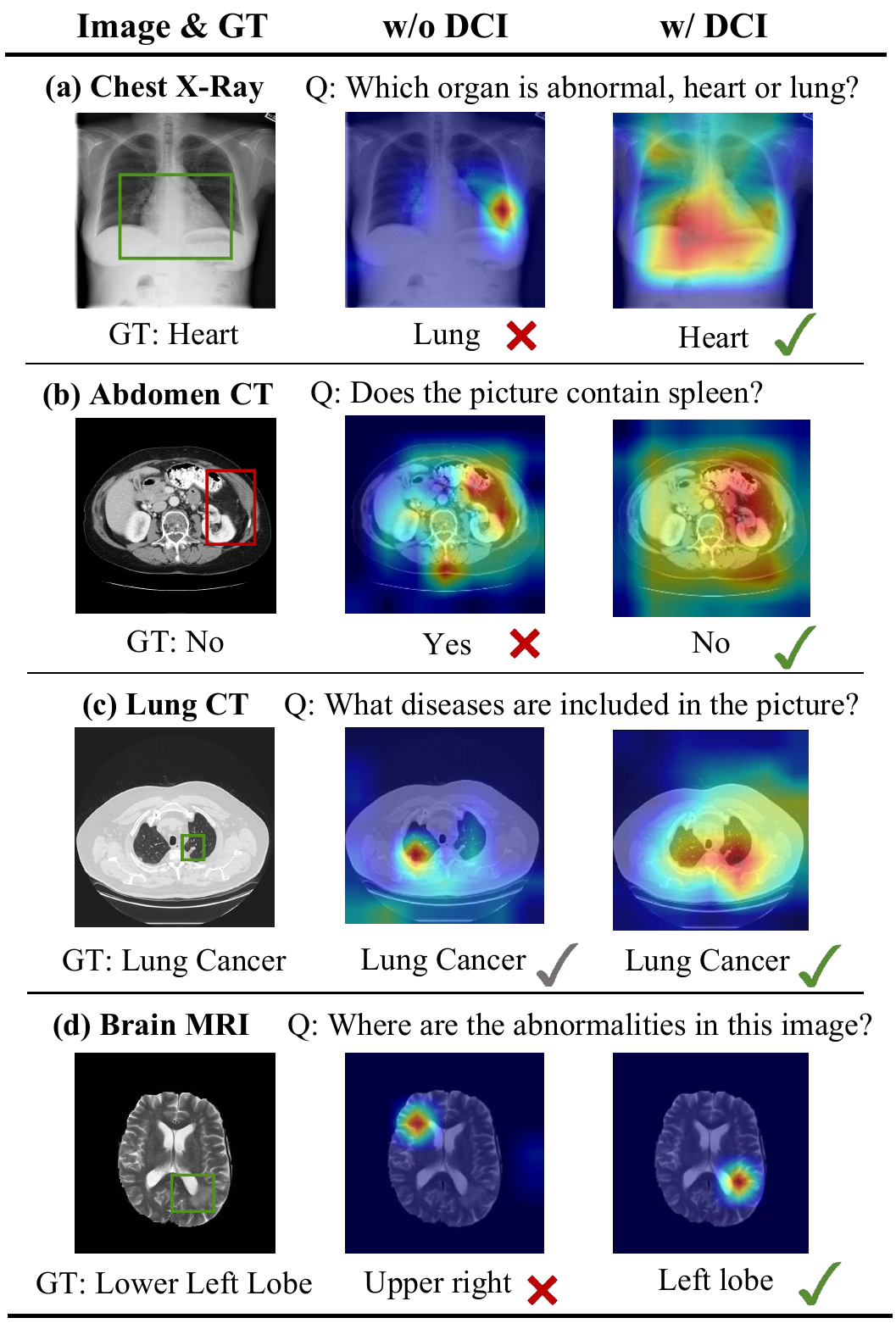}
 \caption{Qualitative results across various medical modalities and organs.
 }
 \label{fig:grad-result}
\end{figure}

\subsubsection{Detailed Ablation on Dictionaries and MI Constraints}

Table~\ref{tab:ablation_detail} provides a fine-grained ablation analysis to evaluate the specific dictionary configurations in the BDA module and individual mutual information (MI) losses in the IV module.

\noindent\textbf{(a) Confounder Dictionaries in BDA:}
We first evaluate the construction of multimodal confounder dictionaries. Replacing our dynamically clustered dictionaries with random initialization drops the overall accuracy noticeably (e.g., from 82.5\% to 76.5\% on VQA-RAD). This result shows that extracting meaningful semantic priors, rather than using arbitrary random vectors, is important for stable causal adjustment. Furthermore, using only the visual ($D_v$) or textual ($D_t$) dictionary leads to partial recovery. The best performance is achieved when both $D_v$ and $D_t$ are used together, confirming that cross-modal biases must be addressed jointly for reliable reasoning.

\noindent\textbf{(b) MI Loss Components in IV:} 
We further analyze the IV module by removing each mutual information (MI) constraint. Removing $\mathcal{L}_{IX}$ causes the most noticeable decline, showing that the \textit{Relevance} condition (linking the IV to the fused representation $X$) is the primary driver for the causal path. More importantly, eliminating $\mathcal{L}_{IC}$ (which enforces \textit{Unconfoundedness} by minimizing MI between $I$ and $C$ for orthogonal disentanglement) or $\mathcal{L}_{IA}$ (which ensures the \textit{Exclusion Restriction} by preventing direct shortcuts to the answer) also leads to clear performance drops. These results align with our theoretical design: all three MI constraints are necessary. Lacking any single term weakens the validity of the instrumental variable, while their combination ensures stable causal representations.

\subsubsection{Sensitivity of Confounder Dictionary Size}

To evaluate the robustness of our causal modules against hyperparameter variations, we analyze the impact of the confounder dictionary size $K \in \{16, 32, 64, 128, 256\}$. Fig.~\ref{fig:dictionary_size_ablation} compares the Full DCI model with its variant lacking the IV module (w/o IV). 

Both configurations exhibit a similar inverted-U trend, achieving optimal performance at $K=64$. A smaller dictionary (e.g., $K=16$) lacks the capacity to capture diverse confounding semantics, while an excessively large dictionary (e.g., $K=256$) introduces redundant noise that interferes with causal adjustment. Crucially, the Full DCI consistently outperforms the ``w/o IV'' variant across all $K$ values. This consistent margin confirms that the performance gain from our IV synthesis is highly robust, effectively mitigating hidden biases regardless of the explicit dictionary capacity.

\subsection{Out-of-Distribution Generalization}
To evaluate whether the model captures invariant causal relations rather than dataset-specific shortcuts, we test it on the Out-of-Distribution (OOD) dataset SLAKE-CP (Table~\ref{tab:ood_sota_comparison}). When shifting from the standard SLAKE to SLAKE-CP, where answer priors are intentionally altered, traditional models like MEVF+BAN suffer severe degradation (dropping to 29.1\% overall accuracy). While existing causal methods like DeCoCT show better robustness (49.2\%), our DCI framework achieves the highest performance across all metrics: 30.2\%, 74.1\%, and 52.9\% in Open, Closed, and Overall accuracy, respectively. Notably, DCI outperforms DeCoCT by a clear margin of 3.7\% overall, with an extremely small variance ($\pm0.4$ to $\pm0.7$) indicating high training stability. These results confirm that our dual causal mechanism successfully prevents the model from overfitting to spurious statistical correlations, ensuring reliable generalization on unseen distributions.

\subsection{Qualitative Analysis}

To visually demonstrate how the proposed DCI framework mitigates confounding biases, we analyze attention maps via Grad-CAM~\cite{selvaraju2017grad} and generated answers across challenging scenarios.

\noindent\textbf{Mitigating Query-Based Hallucination:}
Fig.~\ref{fig:quaresult} illustrates a challenging ``one-image, multi-question'' scenario. The baseline model often locks onto textual keywords (e.g., ``pneumothorax'') and relies on statistical priors rather than dynamic visual reasoning, leading to incorrect localizations and answers. By leveraging the IV mechanism, DCI successfully decouples these entangled hidden confounders. Consequently, the model dynamically shifts its visual attention based on specific query nuances—accurately distinguishing between locating a pneumothorax in the upper right lung and confirming its absence in the lower left (Fig.~\ref{fig:quaresult}(a)). This demonstrates that DCI effectively overcomes query-based hallucinations, ensuring that outputs are grounded in faithful medical reasoning.

\noindent\textbf{Robustness Across Modalities and Organs:}
Fig.~\ref{fig:grad-result} compares attention maps between the baseline (w/o DCI) and our complete model. Without DCI, the model often focuses on visually salient but clinically irrelevant areas due to spurious correlations. For instance, it incorrectly attends to the lung when asked about the heart (Fig.~\ref{fig:grad-result}(a)) and hallucinates a non-existent spleen (Fig.~\ref{fig:grad-result}(b)). In contrast, the DCI-enhanced model produces sharper, concentrated attention maps that align closely with the ground-truth causal regions across various modalities (e.g., Lung CT and Brain MRI). This confirms that the synergy of BDA and IV explicitly marginalizes observable visual-textual co-occurrences and compensates for latent variables, forcing the network to capture invariant clinical semantics rather than superficial features.

\section{Conclusion}
In this paper, we propose a unified framework, \textit{Dual Causal Inference (DCI)}, to systematically address both observed and unobserved confounders in multi-modal MedVQA. DCI integrates two synergistic modules: the Backdoor Adjustment (BDA) module, which marginalizes visible cross-modal spurious correlations via learned dictionaries, and the Instrumental Variable (IV) module, which neutralizes hidden biases. By optimizing with strict mutual information constraints, DCI guarantees the causal disentanglement of multi-modal features, ensuring the model captures genuine medical semantics rather than dataset-specific shortcuts. 
Extensive experiments on three standard benchmarks (VQA-RAD, SLAKE and PathVQA) and an out-of-distribution (OOD) dataset demonstrate that DCI consistently outperforms state-of-the-art methods. Furthermore, qualitative analyses confirm that DCI effectively mitigates query-based hallucinations, promoting clinically meaningful cross-modal attention and generating causally sound, trustworthy diagnostic answers.

\bibliographystyle{IEEEtran}
\bibliography{mybib-2}

\vfill

\end{document}